\documentclass{article}

\usepackage[final, nonatbib]{neurips_2024}

\usepackage[utf8]{inputenc} 
\usepackage[T1]{fontenc}    
\usepackage{url}            
\usepackage[numbers]{natbib}
\usepackage{booktabs}       
\usepackage{amsfonts}       
\usepackage{nicefrac}       
\usepackage{microtype}      
\usepackage{graphicx}
\usepackage{subcaption}
\usepackage{amsmath}
\usepackage{authblk}
\usepackage{hyperref} 
\usepackage{xcolor}
\usepackage{titlesec}

\titlespacing*{\section}{0pt}{+0.3em}{+0.3em}
\pdfoutput=1

\title{Alien Recombination: Exploring Concept Blends Beyond Human Cognitive Availability in Visual Art}

\author[1]{Alejandro Hernandez}
\author[1]{Levin Brinkmann}
\author[1]{Ignacio Serna}
\author[2]{Nasim Rahaman}
\author[3]{Hassan Abu Alhaija}
\author[1]{Hiromu Yakura}
\author[1,4]{Mar Canet Sola}
\author[2]{Bernhard Schölkopf}
\author[1]{Iyad Rahwan}

\affil[1]{\small{Max Planck Institute for Human Development, Berlin, Germany.}}  
\affil[2]{\small{Max Planck Institute for Intelligent Systems, Tübingen, Germany.}}  
\affil[3]{\small{NVIDIA.}} 
\affil[4]{\small{BFM, Tallinn University, Estonia.}}

\begin{document}

\maketitle


\vspace{-3em}
\begin{figure}[ht]
    \centering 
    \includegraphics[width=0.8\textwidth]{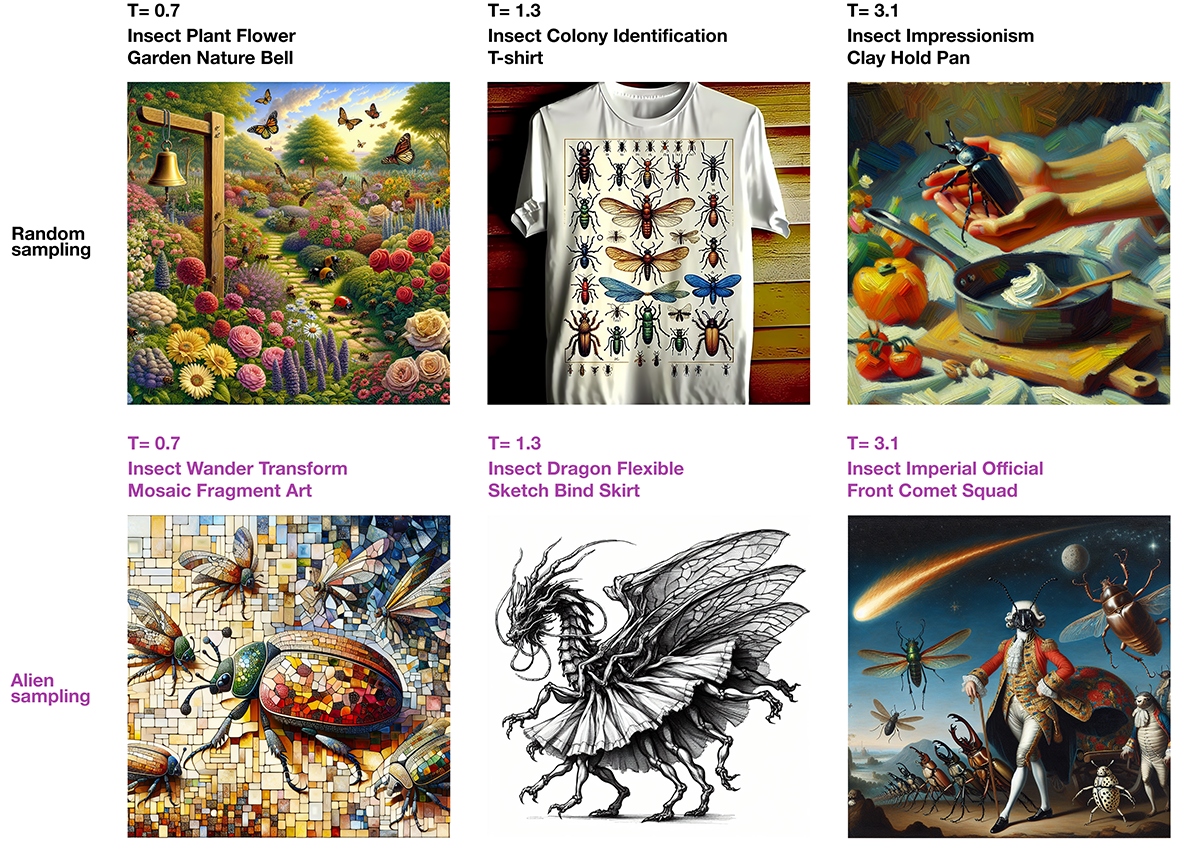} 
    \caption{Comparison of images generated using random sampling versus Alien sampling with the Art model for the input sequence ``Insect''. The Art model, fine-tuned on WikiArt concepts \citep{saleh2015large}, combines likely concepts to produce artworks, with novelty controlled by temperature. Random sampling selects concept combinations arbitrarily, while Alien sampling ranks combinations based on cognitive unavailability and artistic fit, generating images from top-ranked sequences. For the full Alien Recombination method, see Figure \ref{fig:architecture}.} 
    \label{fig:comparison} 
\end{figure}

\vspace{-0.8em}
\section{Introduction}

There is ongoing debate about whether generative AI models can genuinely be considered creative and capable of producing original cultural artifacts. Some AI systems, however, have already broken new intellectual ground. For example, AlphaGo \citep{silver2016mastering} discovered Go strategies that were previously overlooked or unimaginable and have since been learned, adopted, and expanded by human players \citep{superhuman_effects_2023}. Yet, for open-ended domains like art, the question remains: to what extent can these models unlock truly novel and valuable connections that no human mind has conceived?

We hypothesize that in visual art, a vast, unexplored space of concept combinations exists, not due to inherent incompatibility, but because of the limits of the individual artist cultural horizon, including their geographical, temporal, and social embedding. For instance, the concept of ``airplane'' did not exist during the Renaissance period, making it cognitively unavailable to artists of that era. Today, despite familiarity with both concepts, there remains a bias against combining Renaissance style and airplanes. Cognitive science has intensively discussed human availability bias, heavily relying on immediate examples that come to mind when evaluating a specific topic \citep{tversky1973availability}, thereby potentially constraining exploration of novel ideas. Although trained across temporal and cultural boundaries, generative AI models inevitably absorb human biases. As a result, these models can expect to predominantly produce cultural artifacts that align with human cognitive availability. James Evans and his team showed that counteracting such a bias could be key to algorithmic augmented scientific discovery \citep{sourati2023accelerating, shi2023surprising}. Inspired by this work, we developed a system that generates novel visual art concept combinations by modeling and counteracting cognitive biases. It produces combinations not previously attempted in our dataset and cognitively unavailable to any artist in the domain.

\section{Method}

The Alien Recombination method employs two large language models (LLMs) to generate and rank novel combinations of artistic concepts. To build our concept space, we first extracted nine semantic-level features for each image in the WikiArt dataset \citep{saleh2015large} using CLIP \citep{radford2021learning}, selecting the words most similar to the image in CLIP's embedding space. This number was determined through an ablation study showing diminishing returns in concept accuracy beyond nine features. To capture artistic style information, we then added the artwork's style as a tenth concept from metadata, resulting in a comprehensive representation of each artwork.

To facilitate the combinatorial nature of our approach, we constrained concepts to the WordNet Core \citep{boyd2006adding}, a curated list of essential English words. This decision reduces the concept space size, preventing CLIP from extracting multiple words referring to the same concept and increasing concept overlap between artworks. Although this may limit the proper representation of some niche concepts, it enables us to focus purely on novel concept combinations rather than vocabulary variations.

Using these concepts, we created two complementary text datasets: the Art dataset, which contains randomly permuted concept lists for each image, and the Cognitive Availability dataset, created by sampling and aggregating concepts associated with individual artists to reflect their total range of ideas. Let $C = \{c_1, c_2, \dots, c_n\}$ be the set of all possible concepts from WordNet Core. From these datasets, we respectively estimate two key probability distributions: the artwork-level distribution $P_{\text{art}}(c^i | c^0, ..., c^{i-1})$, representing concept co-occurrence within artworks, and the artist-level distribution $P_{\text{cog}}(c^i | c^0, \dots, c^{i-1})$, capturing cognitive availability of concepts to artists, where $c^i \in C \setminus \{c^0, \dots, c^{i-1}\}$ for both distributions.

We then fine-tuned two GPT-2 models \citep{radford2019language} on their respective namesake datasets: the Art model and the Cognitive Availability model. Novel combinations are generated through a two-step process. First, We generate sequences with the Art model, keeping only those with all concepts in WordNet Core. This constraint not only enables later reliable evaluation of combinations within the source data but also ensures the problem remains fully recombinational. Second, we rank these sequences based on perplexity scores from both models. Given that both models accurately represent their target distributions, the perplexity scores provide an approximation of human cognitive biases by capturing which concepts are not usually seen together in artworks or in an artist's full concept usage. Specifically, high perplexity in the Art model indicates concept combinations that rarely co-occur in existing artworks, while high perplexity in the Cognitive Availability model suggests combinations that do not typically appear within an artist's entire body of work, that is, when artists use some concepts in the sequence, they rarely use the others in any of their artworks. In turn, a lower rank in the Art model suggests higher probability of that combination appearing in existing artworks, while a lower rank in the Cognitive Availability model indicates higher cognitive accessibility. 

By inverting the Cognitive Availability rank and using Weighted Rank Aggregation with parameter $\beta$, we obtain concepts that could form coherent artworks while being cognitively unexpected. Higher $\beta$ values emphasize cognitive unavailability, which we term "alienness" as in \citep{sourati2023accelerating}. The top-$k$ ranked sequences from the combined ranking are then returned. We refer to this stage of the method, which involves ranking and selection of sequences, as Alien Sampling. Finally, because the artistic novelty of plain text combinations of concepts can be difficult to evaluate directly, we assess how well these combinations fit in an artwork by visualizing the returned sequences. This is done using a text-to-image model (DALL-E \citep{ramesh2022hierarchical}), with prompts structured as: \texttt{A painting that contains the concepts: <input sequence + generated sequence>}. If a particular style is specified in the sequence, we modify the prompt to reflect that style.

\section{Experiments and results}

We compared the Alien Recombination method, using various values of $\beta$ for Alien sampling, to a baseline method to generate novel images. The Baseline method generates images using the Art model but replaces Alien sampling with random sampling. 

In our experiment, we created 50 unique input sequences, each containing either 1 or 2 concepts. For each input sequence, we generated 150 output sequences at each temperature level, ranging from 0.1 to 3.1 in increments of 0.3, for both the Alien Recombination and Baseline method. From the generated sequences, we then selected the top-ranked sequence for each method based on its respective sampling strategy.

Our experimental design addresses the complex task of assessing artistic novelty through both text-based and image-based evaluation approaches.

\vspace{-0.8em}
\subsection{Text-based novelty}
\vspace{-0.8em}

Our methodology evaluates the novelty of concept combinations using two complementary measures.
\begin{itemize}
        \item \textbf{Novelty relative to artworks:} Let $A = \{A_1, A_2, \dots, A_m\}$ be the set of sets, where each $A_i \subset C$ represents the set of concepts in a single artwork within the Art dataset. Let $S \subset C$ be the set of concepts in the generated sequence. We define the novelty measure $N_{\text{art}}$ as: 
        \begin{equation}
        N_{\text{art}} = \min_{i \in \{1,\dots,m\}} |S \setminus A_i|
        \end{equation}
        This measure represents the minimum number of concepts in $S$ that do not appear together in any single artwork, when compared individually to each artwork in the dataset. A more intuitive way to interpret this measure is that it quantifies the number of distinct concepts in the generated artwork compared to the most similar artwork in the dataset.

        \item \textbf{Novelty relative to cognitive availability:} Let $B = \{B_1, B_2, \dots, B_k\}$ be the set of sets, where each $B_j \subset C$ represents the set of all unique concepts used by artist $j$ in the Art dataset. We define the novelty measure $N_{\text{cog}}$ as:
        \begin{equation}
        N_{\text{cog}} = \min_{j \in \{1,\dots,k\}} |S \setminus B_j|
        \end{equation}
        This measure represents the minimum number of concepts in $S$ that do not appear together in any single artist’s cognitive framework, when compared individually to each artist’s set of concepts. Thus, $N_{\text{cog}}$ quantifies the number of distinct concepts in the generated artwork compared to the most similar artist set.
    \end{itemize}

\begin{figure}[!ht]
    \centering
    \begin{subfigure}[b]{0.45\textwidth}
        \centering
        \includegraphics[width=\textwidth]{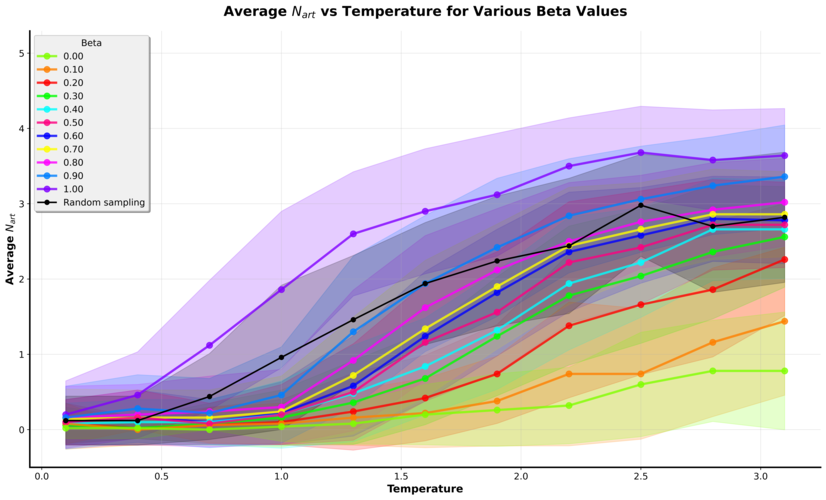}
        \caption{Average $N_{\text{art}}$ vs temperature for multiple values of $\beta$}
        \label{fig:art-dataset-beta}
    \end{subfigure}
    \hfill
    \begin{subfigure}[b]{0.45\textwidth}
        \centering
        \includegraphics[width=\textwidth]{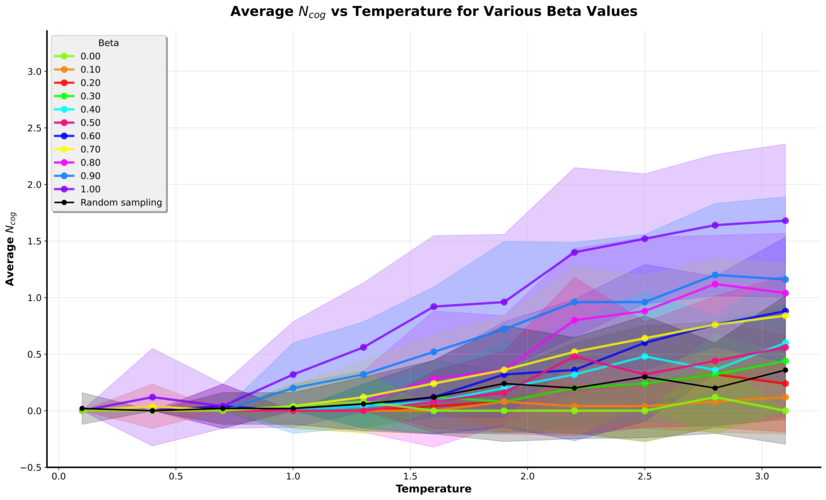}
        \caption{Average $N_{\text{cog}}$ vs temperature for multiple values of $\beta$}
        \label{fig:cog-dataset-beta}
    \end{subfigure}
    \caption{Comparison on the novelty of the generated sequences with respect to the artworks and the cognitive availability for multiple values of $\beta$}
    \label{fig:comparison-dataset-beta}
\end{figure}
    
Our findings reveal that while increasing the Art model's temperature can generate novel combinations absent from the dataset at artwork-level (Figure \ref{fig:art-dataset-beta}), it does not reliably produce cognitively unavailable combinations. The Alien Recombination method, through its explicit search for cognitive unavailability, demonstrates a consistently higher likelihood of generating such combinations (Figure \ref{fig:cog-dataset-beta}). Empirically, we show that generating unseen combinations in artworks when increasing the temperature is relatively straightforward, with 85\% of the combinations containing at least one new concept when surpassing temperature 1. However, this phenomenon does not extend to cognitively unavailable combinations: at temperature 3, more than half of the combinations have $N_{\text{cog}}= 0$, indicating no cognitively unavailable concepts. This reveals that finding completely cognitively unavailable combinations is a fundamentally much harder task, requiring explicit search strategies within the concept space. Further analysis and details are provided in the Appendix.

\vspace{-0.8em}
\subsection{Image-based Novelty}
\vspace{-0.8em}

For both the Baseline method and the Alien Recombination method with $\beta = 0.85$, we generated sequences and converted the top-ranked sequences into images for evaluation. Image novelty was then evaluated using two independent approaches:

\begin{itemize}
    \item \textbf{Evaluation using GPT-4:} We used GPT-4 \citep{achiam2023gpt} to perform pairwise comparisons between images produced by the Alien Recombination and Baseline methods. Based on GPT-4's alignment with human evaluators in vision-language tasks \citep{zhang2023gpt}, we asked it to evaluate concept combination novelty. For each pair of images created with the same input and temperature, GPT-4 was prompted with the following request: \texttt{As an art expert, please write a sentence indicating which image is more novel, focusing on concept combination novelty.}

    \item \textbf{Embedding-based Analysis:} Following the approach of \citep{elgammal2018shape}, we computed ResNet152 \citep{he2016deep} embeddings for both the generated images and those in the WikiArt dataset. Then, we assessed image novelty in two ways: (1) For each generated image, we computed its maximum cosine similarity with any WikiArt image, where lower maximum similarity indicates higher novelty. We then compared the average of these maximum similarities between methods. (2) As before, we performed pairwise comparisons between images generated by both methods, considering the image with lower similarity to be more novel.
\end{itemize}

\begin{figure}[!ht]
    \centering 
    \includegraphics[width=0.5\textwidth]{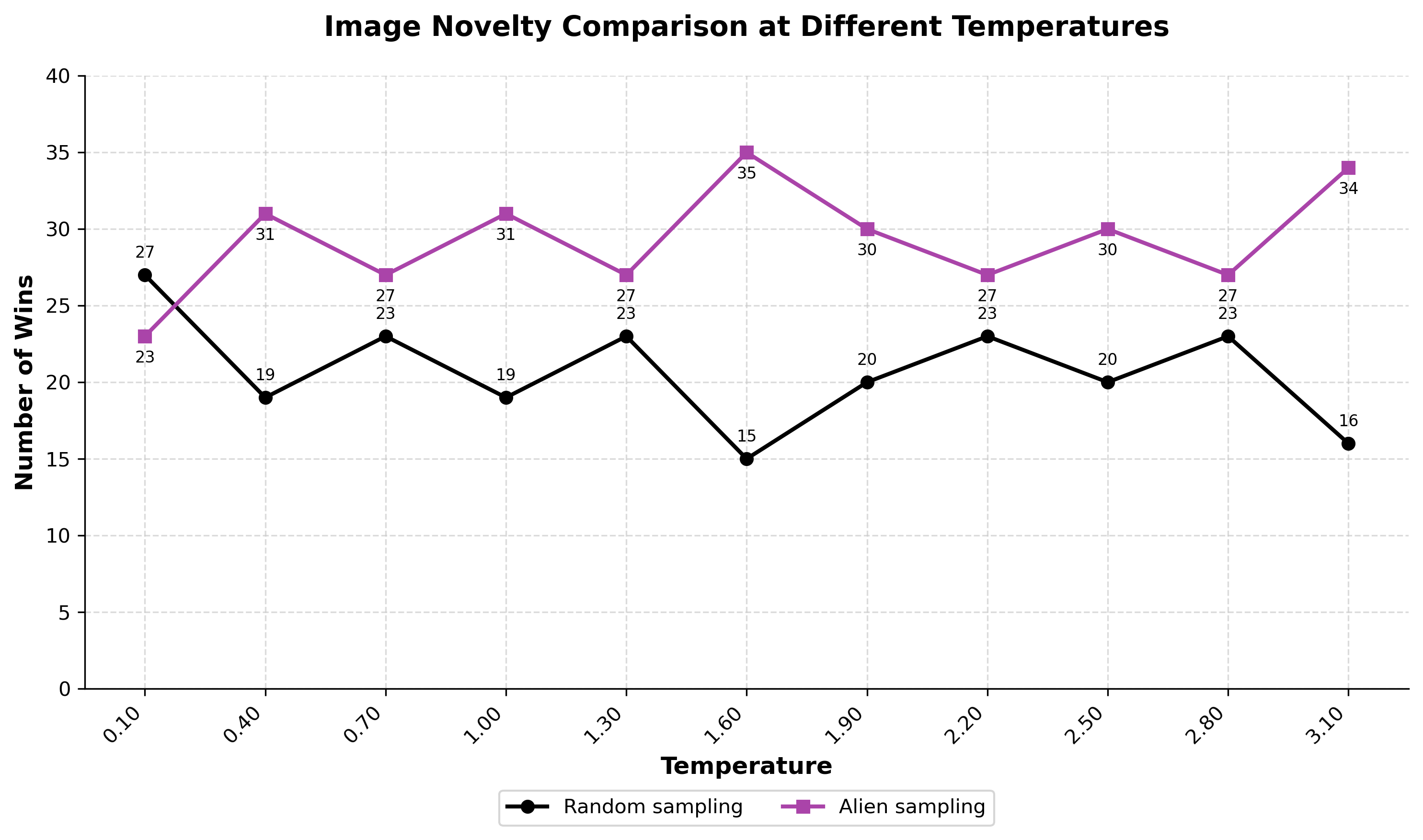}
    \caption{GPT-4 pairwise evaluation} 
    \label{fig:comparison-novelty} 
\end{figure}

GPT-4 evaluations consistently rated Alien Recombination images as more novel on average, suggesting that cognitively unavailable combinations are perceived as more innovative (Figure \ref{fig:comparison-novelty}). The embedding analysis corroborated these findings, showing lower average similarity scores and more favorable pairwise comparisons for the Alien Recombination method (Figure \ref{fig:embedding-analysis}).

However, it is important to note that ResNet embeddings also account for visual factors such as lighting and contrast, making them sensitive to elements beyond just conceptual differences. In contrast, GPT-4 can be explicitly prompted to evaluate novelty based on conceptual combinations. Additionally, since DALL-E generates in-distribution images, its outputs may be constrained by familiar patterns and styles, potentially limiting its ability to fully capture the novelty of the prompt. Further details can be found in the Appendix.

\section{Conclusion}
We present the Alien Recombination method, designed to generate novel artistic combinations within the space of visual art. This system not only produces combinations that have never been attempted before within our dataset but also identifies and generates combinations that are cognitively unavailable to all artists in the domain. Additionally, our results suggest that cognitive unavailability is a promising metric for optimizing artistic novelty, outperforming mere temperature scaling. This approach uses generative models to connect previously unconnected ideas, providing new insight into the potential of framing AI-driven creativity as a combinatorial problem \citep{thagard2011aha, gero1996creativity}.

\bibliographystyle{unsrt}  
\bibliography{references}

\section*{Appendix}

\subsection*{Architecture Overview}

\begin{figure}[ht]
    \centering 
    \includegraphics[width=1\textwidth]{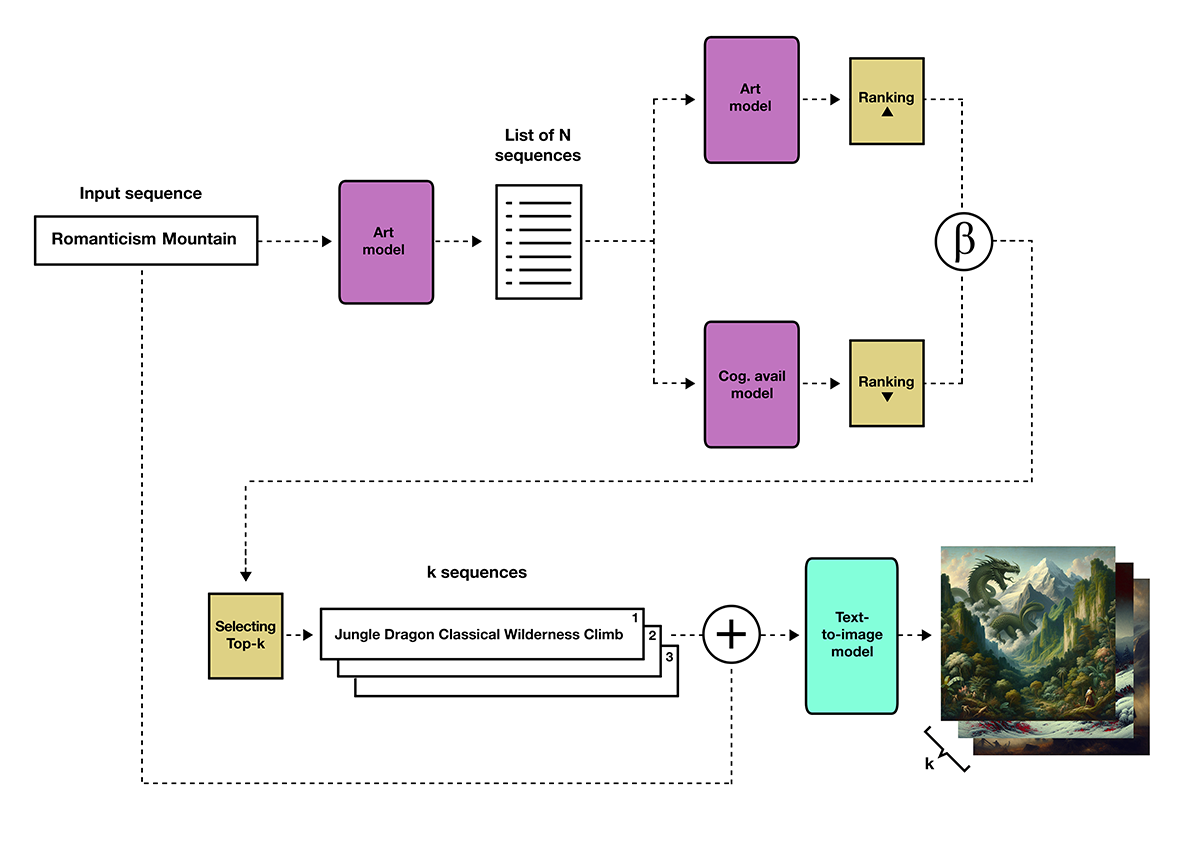} 
    \caption{Schematic representation of the Alien Recombination method. The Art model generates N sequences from an input sequence. These sequences are ranked by both the Art model (ascending perplexity) and the Cognitive Availability model (descending perplexity). The ranking and selection process, termed ``Alien sampling'', employs a weighted rank aggregation method parameterized by \(\beta\). Increasing \(\beta\) prioritizes sequences that are more distant from what is cognitively available, thus enhancing \textit{alieness}. The top-$k$ sequences (user-defined $k$) resulting from this fused ranking are then processed by a text-to-image model (DALL-E \citep{ramesh2022hierarchical} in this study) to generate images, using the prompt: \,\texttt{A painting that contains the concepts:\,<input sequence + generated sequence>}.}
    \label{fig:architecture} 
\end{figure}

The complete workflow of the Alien Recombination method is depicted in Figure \ref{fig:architecture}.
 
\subsection*{Experiment: Multiple images for the same input sequence}

We tested the Alien Recombination method with \(\beta = 0.7\) by generating various images from the same input sequence across a range of temperature values. Increasing the temperature results in more novel and diverse sequences from the Art Model providing the Alien sampling with a richer range of sequences to evaluate, allowing it to identify and return those that are more distant from what is cognitively available.

\begin{figure}[!ht]
    \centering 
    \includegraphics[width=1\textwidth]{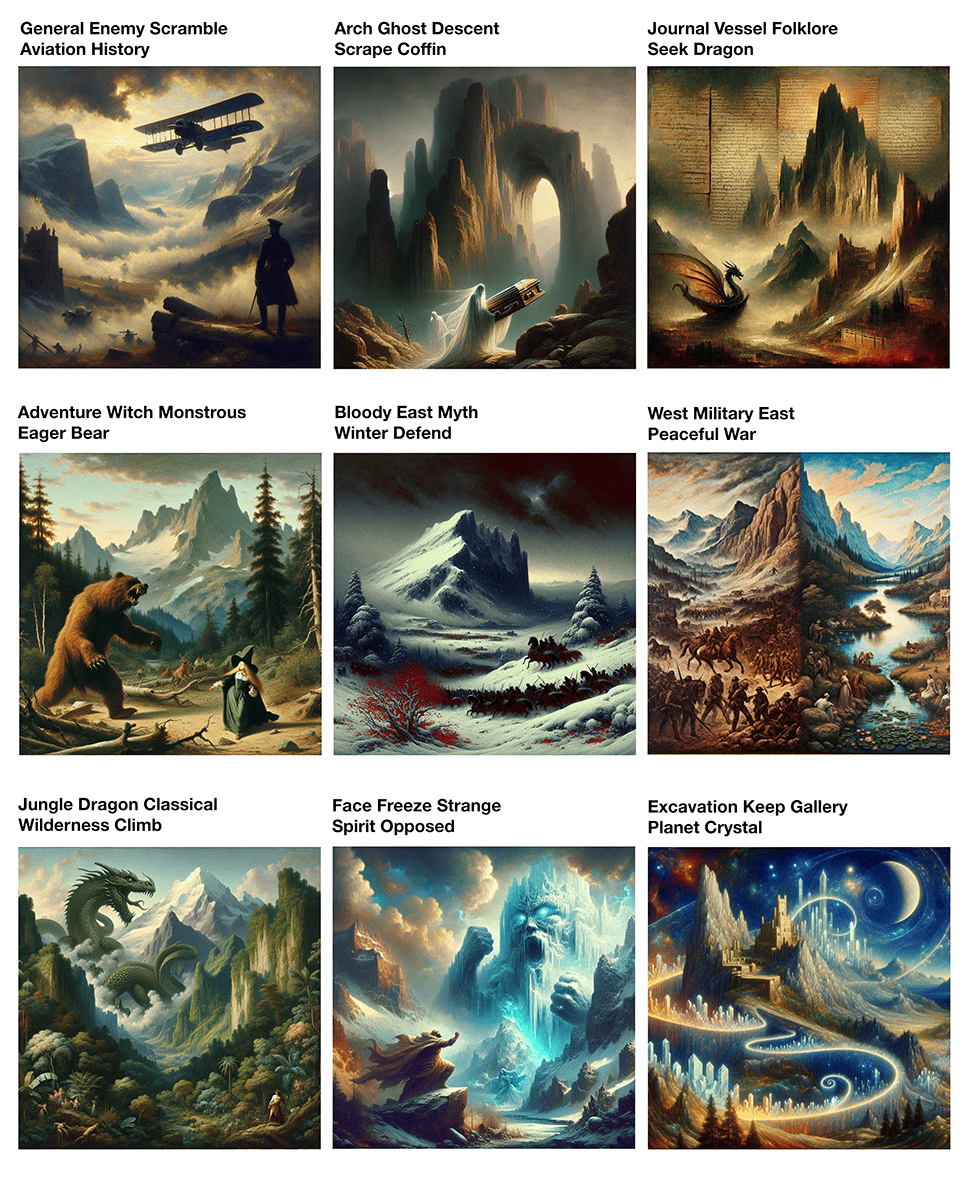}
    \caption{Examples of images generated with the Alien Recombination method for the input ``Romanticism Mountain''. The returned sequence by the Alien sampling is written on top of the respective image.} 
    \label{fig:romanticism-mountain} 
\end{figure}

\begin{figure}[ht]
    \centering
    \begin{subfigure}[b]{1\textwidth}
        \centering
        \includegraphics[width=\textwidth]{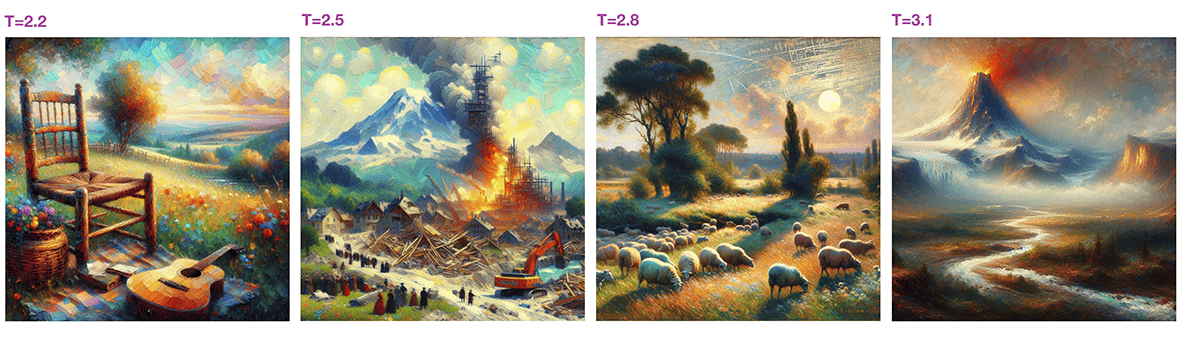}
        \caption{Generated images by the Alien Recombination method given the input ``Impressionism Landscape''}
        \label{fig:impressionism}
    \end{subfigure}
    \hfill
    \begin{subfigure}[b]{1\textwidth}
        \centering
        \includegraphics[width=\textwidth]{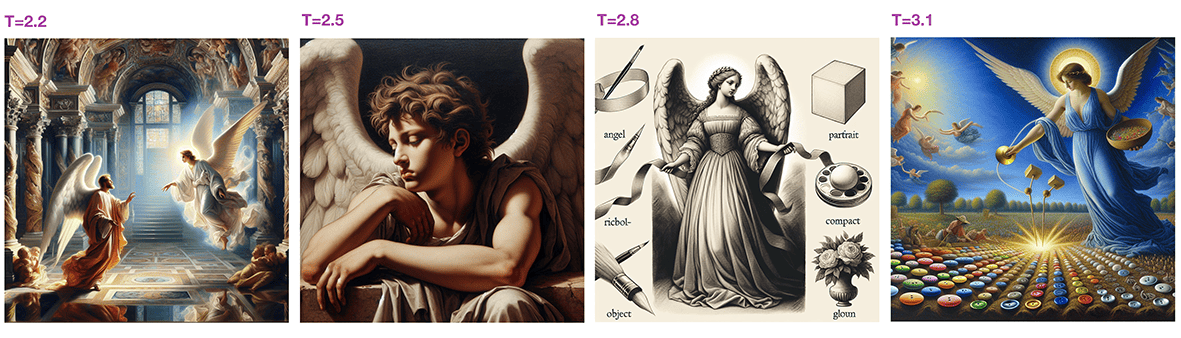}
        \caption{Generated images by the Alien Recombination method given the input ``High-Renaissance Angel''}
        \label{fig:angel}
    \end{subfigure}
    \caption{Generated images by the Alien model for different style inputs}
    \label{fig:alien-more-styles}
\end{figure}

Figure \ref{fig:romanticism-mountain} illustrates examples of images generated by the Alien model, using ``Romanticism Mountain'' as input sequence. Each example includes the sequence returned by the Alien sampling. This returned sequence, along with the input sequence, is subsequently used in the text-to-image model to produce the final images. Figure \ref{fig:alien-more-styles} shows more examples of introducing other styles as input sequence.

It is important to note that in this study the temperature parameter does not influence the model in the same way as it does in conventional language models. Using a very high temperature with the Art Model can result in sequences that include non-existent or incomplete tokens, which cannot form meaningful artworks. To address this issue, we force the generated sequences to be composed only of words from WordNet Core, the vocabulary used to extract concepts from the artworks. This approach prevents us from introducing words that do not belong to the vocabulary of concepts we are using. In this way, we preserve the problem as pure recombinational. Therefore, we continue generating sequences until the desired number of valid sequences is obtained. This method ensures that we maintain more meaningful and stable outputs even at high temperatures, such as 3.0, unlike conventional language models used for general purposes.

\clearpage
\subsubsection*{Artwork-Level Novelty is Easy, Artist-Level Novelty is Hard}
\begin{figure}[!htbp]
    \centering
    \begin{subfigure}[b]{0.48\textwidth}
        \centering
        \includegraphics[width=\textwidth]{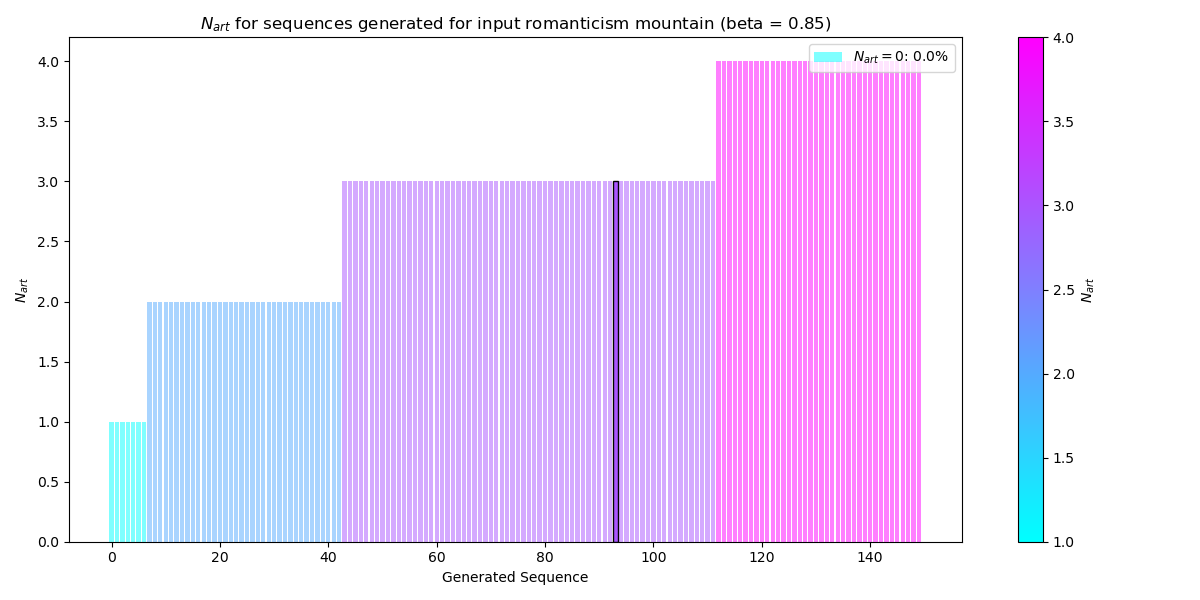}
        \caption{$N_{\text{art}}$ with $\beta = 0.85$.}
    \end{subfigure}
    \hspace{0.5mm} 
    \begin{subfigure}[b]{0.48\textwidth}
        \centering
        \includegraphics[width=\textwidth]{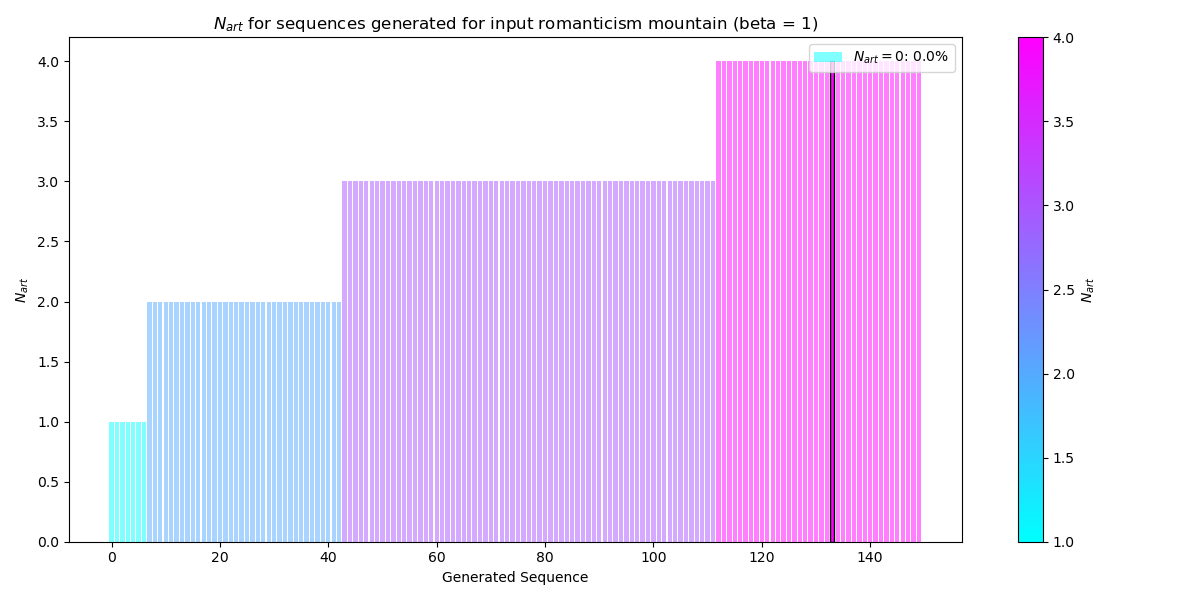}
        \caption{$N_{\text{art}}$ with $\beta = 1$.}
    \end{subfigure}
    \begin{subfigure}[b]{0.48\textwidth}
        \centering
        \includegraphics[width=\textwidth]{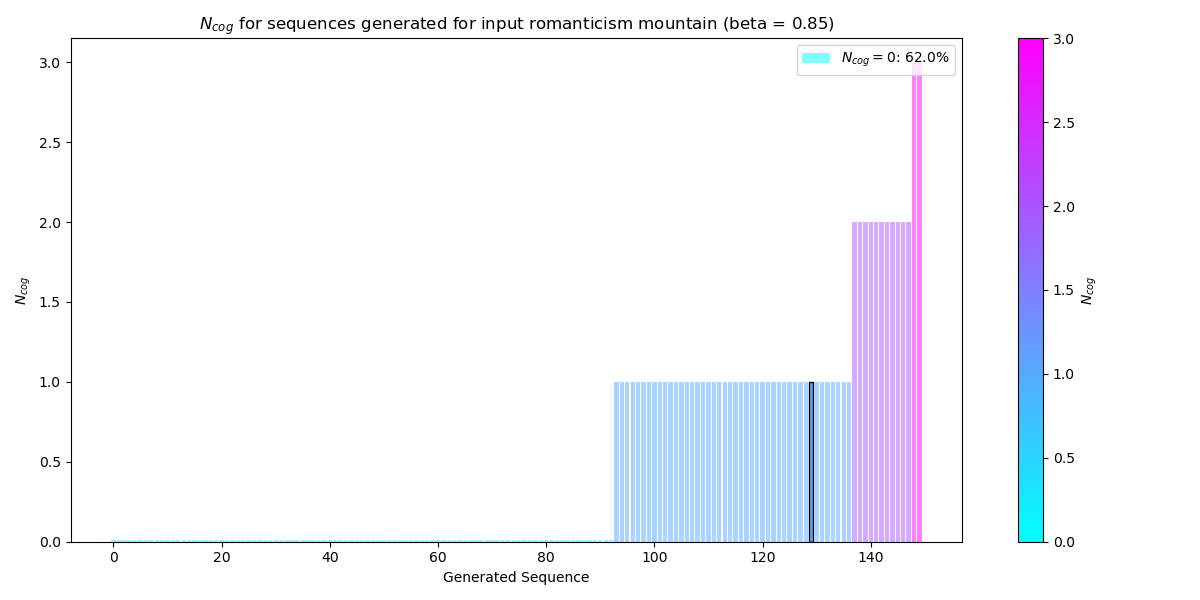}
        \caption{$N_{\text{cog}}$ with $\beta = 0.85$.}
    \end{subfigure}
    \hspace{0.5mm} 
    \begin{subfigure}[b]{0.48\textwidth}
        \centering
        \includegraphics[width=\textwidth]{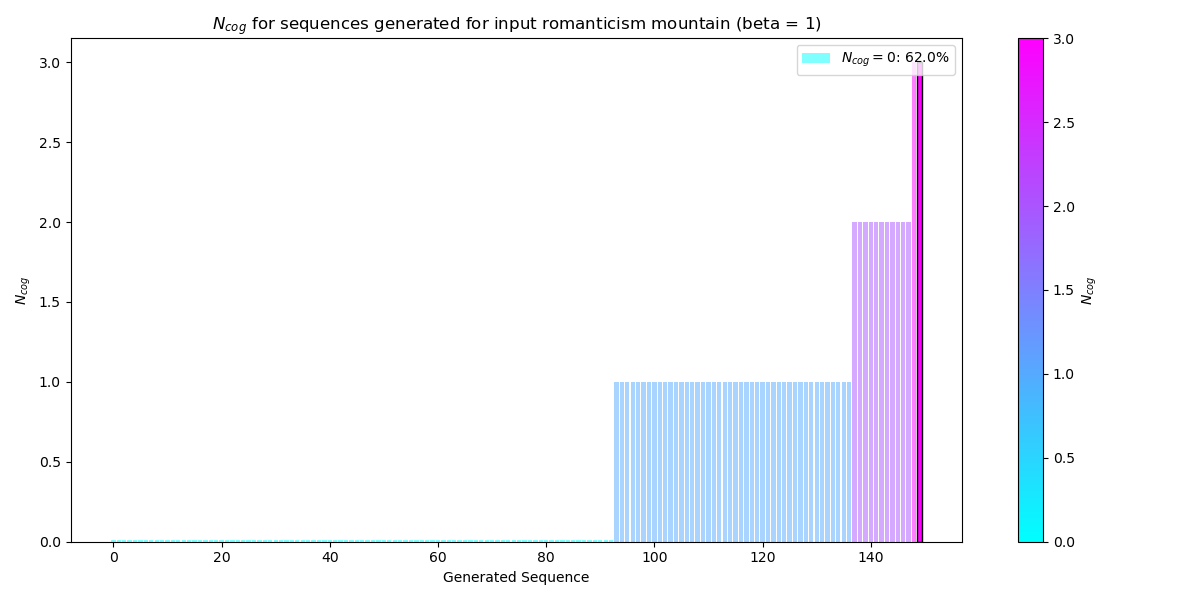}
        \caption{$N_{\text{cog}}$ with $\beta = 1$.}
    \end{subfigure} 
    \caption{Comparison between selected sequences with Alien samplig with $\beta = 0.85$ and $\beta = 1$ for input "Romanticism Mountain" and temperature 2.5. The sequences are sorted by $N_{\text{art}}$ and $N_{\text{cog}}$, depending on the measure plotted. The selected sequence is highlighted in black. The proportion of not cognitive unavailable ($N_{\text{cog}} = 0$) sequences is in the right corner.}
    \label{fig:comparison-beta-n}
\end{figure}

We plotted \(N_{\text{art}}\) and \(N_{\text{cog}}\) for each sequence generated by the Art model across various temperature levels, highlighting the selected sequence using Alien sampling with $\beta = 0.85$ (Figure \ref{fig:comparison-beta-n}). The results clearly show that, even at high temperatures, finding cognitively unavailable sequences is much harder than finding artistically novel ones. For example, at a temperature of 2.5, all generated sequences have \(N_{\text{art}} \geq 1\), with many sequences showing \(N_{\text{art}} \geq 3\). However, 62.0\% of these sequences are not cognitively unavailable (\(N_{\text{cog}} = 0\)), with a notable difference in density. This highlights the need for a method that explicitly detects and selects cognitively unavailable combinations, as merely temperature scaling has proven to be an inefficient approach.

Finally, the impact of \(\beta\) on sequence selection is evident in Figure \ref{fig:comparison-beta-n}. For the same input and temperature level, \(\beta = 1\) typically selects the sequence with the highest \(N_{\text{cog}}\). In contrast, \(\beta = 0.85\) occasionally selects sequences with lower \(N_{\text{cog}}\), reflecting the trade-off between maximizing cognitive unavailability and favoring sequences that are more plausible in the context of existing artworks.

\clearpage
\subsubsection*{GPT-4 image evaluation}
\begin{figure}[!ht]
    \centering
    \begin{subfigure}[b]{1\textwidth}
        \centering
        \includegraphics[width=\textwidth]{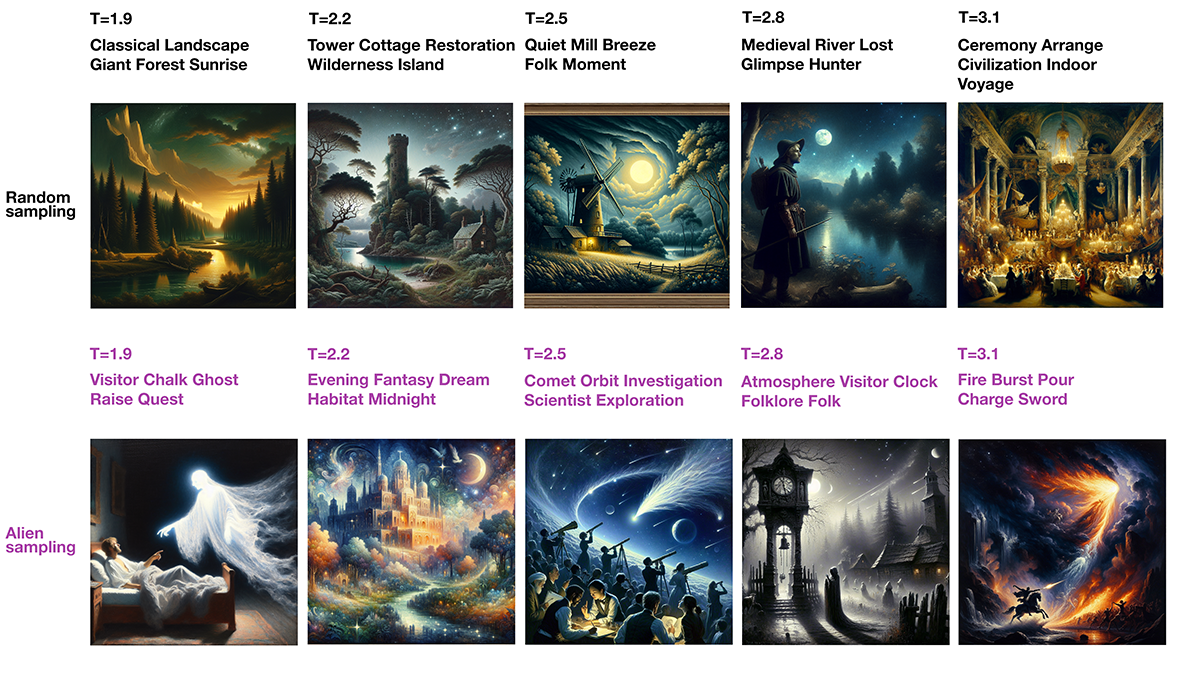}
        \caption{Example of sequences and images generated for the input "Romanticism Night" using the baseline method (random sampling) and the Alien Recombination method (Alien sampling).}
        \label{fig:romanticism-night}
    \end{subfigure}
    \hfill
    \begin{subfigure}[b]{0.8\textwidth}
        \centering
        \includegraphics[width=\textwidth]{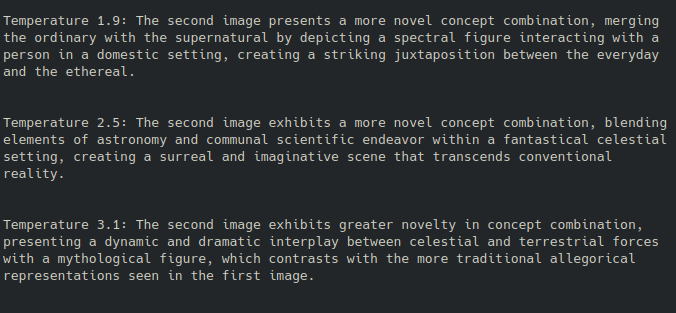}
        \caption{Example of the GPT-4 output for the comparison.}
        \label{fig:gpt4-comp}
    \end{subfigure}
    \caption{Results for the Baseline and Alien Recombination methods for the input "Romanticism Night", along with comparisons provided by GPT-4.}
    \label{fig:romanticism-night-comp-gpt4}
\end{figure}

To evaluate the novelty of the images generated from the sequences selected by both sampling methods, we used GPT-4 \citep{achiam2023gpt}. Given that GPT-4 has been shown to align closely with human evaluators in general vision-language tasks \citep{zhang2023gpt}, we employed it to determine which image is more novel. For each temperature and input, we compared the images generated using a random sampling, the Baseline method, and the Alien sampling. We prompted GPT-4 with the instruction: \texttt{As an art expert, please write a sentence indicating which image is more novel, focusing on concept combination novelty}, and selected the more novel image based on GPT-4's response (Figure \ref{fig:romanticism-night-comp-gpt4}). This process revealed that the images generated by the Alien sampling consistently exhibited greater novelty than those produced by the random sampling for the same input and level of temperature, as illustrated in Figure \ref{fig:comparison-novelty}.

\subsubsection*{Embedding evaluation}
To further evaluate the novelty of our generated images, we drew inspiration from \citep{elgammal2018shape}. We employed a ResNet architecture \citep{he2016deep} to embed all images in WikiArt. This approach has been shown to internally represent and
organize artworks chronologically and stylistically without prior knowledge of art history, while also potentially identifying influential artists based on the positioning of their works’ embeddings within style clusters.

\begin{figure}[ht]
    \centering 
    \includegraphics[width=0.7\textwidth]{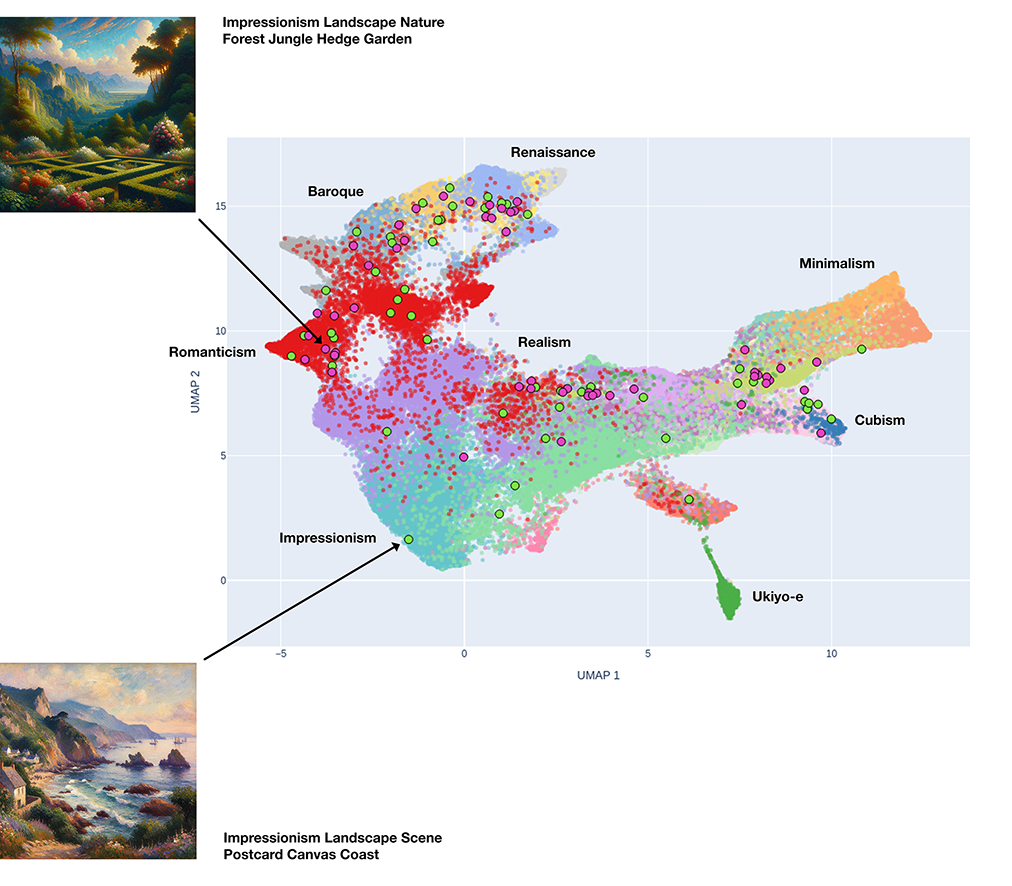}
    \caption{2D embeddings of WikiArt images and generated images for temperature
1.0 for the Baseline (lime) and Alien Recombination (pink) methods in
UMAP space.} 
    \label{fig:emb-space} 
\end{figure}

We fine-tuned a ResNet with 152 convolutional layers and 1 linear layer for WikiArt style classification. Our model achieved 59.7\% accuracy on the validation set, comparable to the 60.2\% reported in the original study. As illustrated in Figure \ref{fig:emb-space}, the WikiArt styles arrange chronologically, confirming the findings of Elgammal et al. Notably, the Ukiyo-e style stands out, distinctly separated from other styles, as it is the only
non-Western style in the dataset. This clear separation indicates that ResNet’s ability to capture chronological patterns is linked to its ability to trace the flow of artistic influence across movements.

For each generated image in both methods, we computed the closest cosine similarity to the embedding of an image in the WikiArt dataset. We then compared the average closest cosine similarity for each method and temperature level.

\begin{figure}[ht]
    \centering 
    \includegraphics[width=0.5\textwidth]{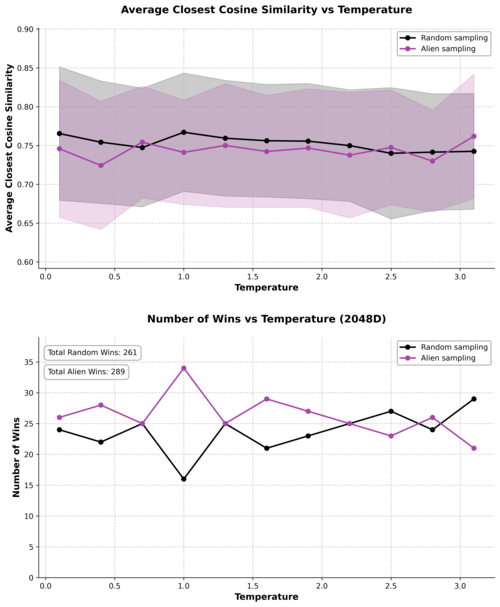}
    \caption{Average cosine similarity and number of wins vs Temperature} 
    \label{fig:embedding-analysis} 
\end{figure}

In addition, to determine which method produces more novel images for the same input and temperature level, we conducted a pairwise comparison. For each pair of generated images with the same input and temperature, we designated the image with lower cosine similarity as more novel. The results are depicted in Figure \ref{fig:embedding-analysis}

As Elgammal et al. noted, neural networks like ResNet can learn various features to discriminate between styles, including compositional elements, contrast, color composition, brush strokes, and subject matter. However, our use of DALL-E to generate images introduces numerous variables in composition that we cannot control, all of which influence the similarity assessment. Moreover, DALL-E is designed to generate images that appear reasonable and within distribution, potentially constraining the extent of visual novelty it can produce.

This limitation may make our embedding-based evaluation less stable for assessing
novelty in concept combinations, as it captures a broader range of visual features beyond just conceptual novelty. In contrast, GPT-4 can be explicitly prompted to focus on conceptual combinations, providing a more targeted evaluation of our specific research interest.

Additionally, DALL-E demonstrates proficiency in replicating specified styles, as
evidenced by many images embeddings appropriately within their respective style
cluster. However, focusing on Figure \ref{fig:emb-space}, this experiment also reveals some limitations of DALL-E and this type of evaluation.

For instance, for one of the generated inputs, "Impressionism Landscape", the random sampling method selected a sequence containing concepts more common in that context: "Scene Postcard Canvas Coast". This sequence had perplexity scores of 9.24 for the Art model and 25.03 for the Cognitive Availability model.
In contrast, the Alien Recombination method generated the sequence "Nature Forest
Jungle Hedge Garden". Despite having $N_{\text{cog}} = 0$, this sequence showed higher perplexity scores: 12.38 for the Art model and 58.69 for the Cognitive Availability model, indicating that this combination of concepts is less likely to be cognitively available.

When generating images, DALL-E correctly applied the Impressionist style to the
Baseline method’s input. However, for the Alien Recombination input, it produced
a more Romantic-style image, as indicated by the position of the embedding. This
discrepancy could come from the unusual combination of "Impressionism" and "Jungle", along the almost opposite concept of "Garden" (as "Jungle" and "Garden" usually represent highly different types of vegetation and structure) for which there are less examples. DALL-E, trained primarily on in-distribution images, defaulted to a more familiar Romantic interpretation of a jungle scene.

This example illustrates how the novel combinations produced by the Alien Recombination method can challenge DALL-E’s accuracy, especially when given short, non-informative prompts like we use in this study. It shows DALL-E’s tendency to revert to more familiar representations when faced with unusual concept combinations, potentially adding more noise to this already noisy evaluation.

\subsection*{Ethical Implications}
This method can serve as a tool for discovering new connections in art that might be overlooked due to cultural, temporal and spatial gaps. However, the Alien Recombination method currently proposes combinations based on its training dataset, which may be biased towards 20th century Western visual art. Additionally, notions of appeal and novelty can vary depending on an individual’s background, so the model’s optimization for appeal and novelty might be skewed towards the preferences of the dataset’s demographic.

\begin{figure}[ht]
    \centering 
    \includegraphics[width=1\textwidth]{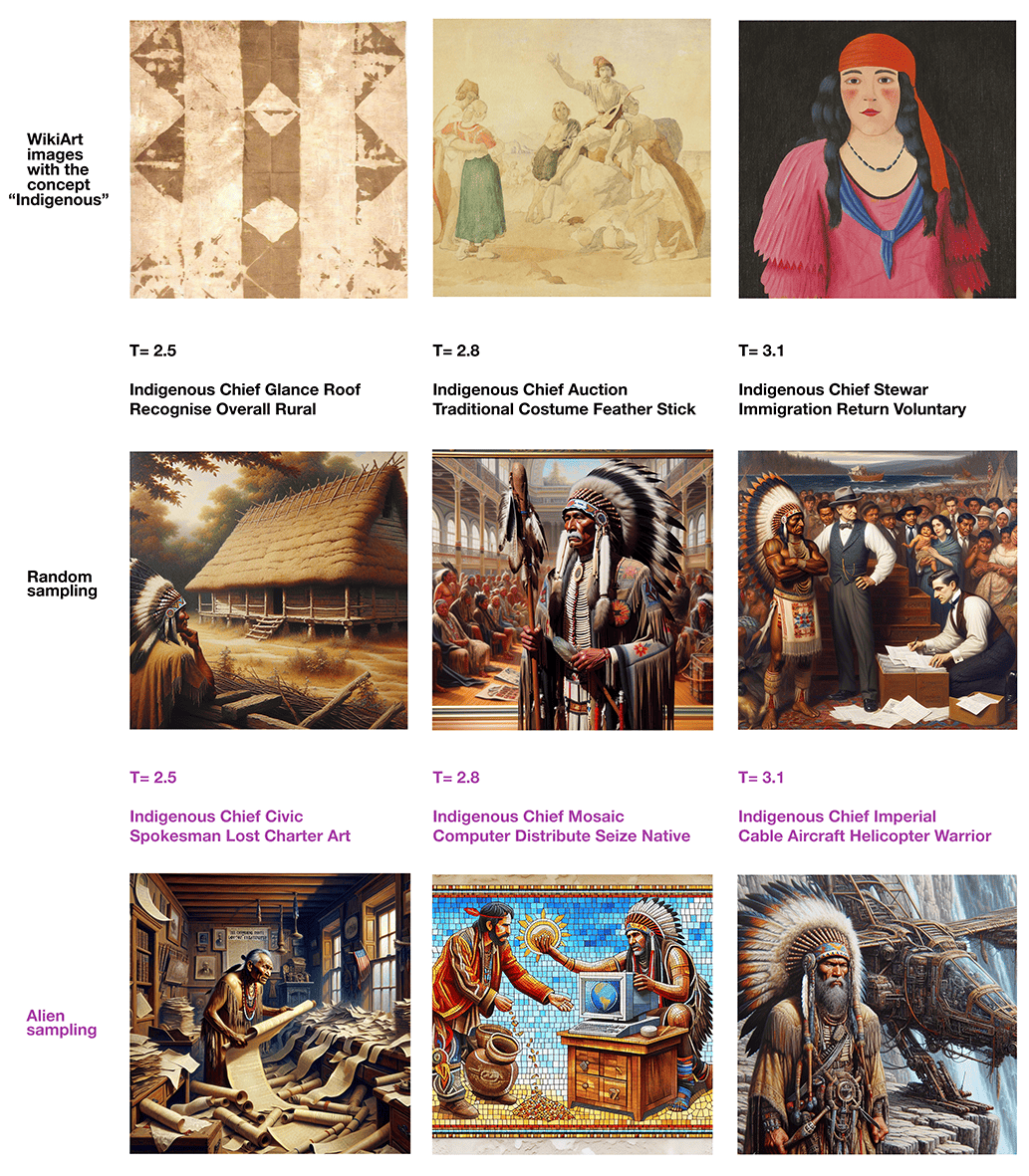} 
    \caption{Comparison of images that contains the concept ``Indigenous'' in the WikiArt dataset with the images generated using the random sampling and Alien sampling.} 
    \label{fig:indigenous} 
\end{figure}

It is crucial to acknowledge that our method inherits the biases present in its constituent models, CLIP and DALL-E. During our experiment with multiple inputs, one of the generated inputs was the sequence ``Indigenous Chief''. As illustrated in Figure \ref{fig:indigenous}, CLIP's interpretation of ``Indigenous'' manifested through a diverse range of paintings, reflecting its own biases in concept recognition. Conversely, DALL-E depicted the concept ``Indigenous'' predominantly through a narrow, stereotypical lens of Native American imagery.

In addition to inheriting biases from the models composing the Alien Recombination method, it is essential to understand that the notion of cognitive availability is also biased. The method’s perception of what is cognitively available is heavily influenced by its training data, which is based on what is depicted in paintings and may not accurately represent reality. Consequently, the method might produce combinations that are categorized as cognitively unavailable, such as ``Indigenous Computer'' shown in Figure \ref{fig:indigenous}. Although these combinations may be underrepresented or missing in the dataset, they are entirely plausible and relevant in today's society. This highlights the risk that concepts currently recognized in society may be misrepresented or outdated in the dataset due to biases in the training data.

\end{document}